\documentclass{article} 
\usepackage{iclr2027_conference,times}

\usepackage{amsmath,amsfonts,bm}

\def\eqref#1{equation~\ref{#1}}

\def\1{\bm{1}}

\DeclareMathAlphabet{\mathsfit}{\encodingdefault}{\sfdefault}{m}{sl}
\SetMathAlphabet{\mathsfit}{bold}{\encodingdefault}{\sfdefault}{bx}{n}

\usepackage{hyperref}
\usepackage{url}

\usepackage{graphicx}
\usepackage{subcaption}
\usepackage{wrapfig}
\usepackage{tabularx}
\usepackage{booktabs}
\usepackage{multirow}
\usepackage{xcolor}
\usepackage{colortbl}
\usepackage{bm}
\usepackage{subcaption}
\usepackage{enumitem}
\usepackage{amssymb}

\newcommand{\eg}{\textit{e}.\textit{g}., }

\title{Urgent Actions Go First: Urgency-Aware Denoising for Real-Time VLA Control}

\author{%
Zibo Wang \\
    Pengcheng Laboratory \\
    \And
    Haochen Han \\
    Pengcheng Laboratory \\
    \And
    Pengzhen Ren \\
    Pengcheng Laboratory \\
    \And
    Mingtong Dai \\
    Pengcheng Laboratory \\
    \And
    Fangming Liu \\
    Pengcheng Laboratory \\
}

\iclrfinalcopy 
\begin{document}

\maketitle
\fancyhead[L]{Preprint}

\begin{abstract}
Diffusion and flow-matching Vision-Language-Action (VLA) policies generate action chunks through iterative denoising, incurring substantial inference latency that severely limits real-time robotic control. Existing acceleration methods treat an action chunk as a monolithic computational unit, ignoring a crucial physical reality of receding-horizon control: actions are generated jointly but consumed sequentially, resulting in inherently heterogeneous execution urgencies. We exploit this asymmetry to introduce Urgency-Aware Denoising (UAD), a novel inference-time framework that allocates denoising computation according to when each action is physically needed. UAD releases time-critical urgent actions after fewer denoising steps while overlapping the continued background refinement of tail actions with physical execution. However, heterogeneous denoising introduces two key challenges: early-release errors in urgent actions and trajectory inconsistency in tail actions. UAD elegantly resolves both through two core mechanisms: Trajectory Reconciliation, which reconstructs unified internal state evolution to restore joint denoising coherence without additional model evaluations, and Ghost Action Correction, which leverages non-executed ghost continuations to dynamically compensate for early-release errors across remaining executable actions. Extensive evaluations across multiple VLA architectures, simulation benchmarks, and real-world manipulation tasks demonstrate that UAD achieves up to a $1.89\times$ speedup in average action availability latency while maintaining comparable success rates to vanilla inference with optimal denoising budget, offering a more favorable success-latency trade-off than state-of-the-art VLA acceleration baselines.

\end{abstract}

\section{Introduction}

Diffusion and flow-matching policies have emerged as a dominant paradigm for Vision-Language-Action (VLA) models, enabling expressive and high-quality generation of temporally coherent action sequences \citep{chi2025diffusion,black2024pi_0,intelligence2025pi_}. In diffusion-based VLA policies, actions are generated in chunks, where each chunk is progressively refined through iterative denoising steps \citep{intelligence2025pi_}. While this iterative process improves action quality, repeated evaluations of the diffusion transformer (DiT) incur substantial inference latency, which severely constrains real-time robotic control \citep{black2026real}. In a typical setting of our measurements, executing an action chunk takes 500~ms, while generating the next chunk requires approximately 300~ms even on datacenter-grade hardware. Under synchronous execution, this chunk-level generation delay forces the robot to spend over 30\% of each execution cycle idle while waiting for new actions, a bottleneck that is more pronounced on resource-constrained edge devices \citep{nguyen2026vla}.

To mitigate this inference bottleneck, existing acceleration methods have explored general optimizations (\eg quantization \citep{xu2026qvla}, pruning \citep{pei2026action}, and feature caching \citep{ji2026block}) as well as VLA-oriented techniques like adaptive step reduction \citep{yu2025d3p} and generation-execution overlapping \citep{black2026real}. However, existing acceleration approaches largely treat an action chunk as a monolithic computational unit, applying a shared denoising budget across all action tokens. This overlooks a fundamental physical reality in receding-horizon control: \textit{actions within a chunk are generated jointly but consumed sequentially, resulting in inherently heterogeneous execution urgencies.}

\begin{figure}
    \centering
    \includegraphics[width=1.0\linewidth]{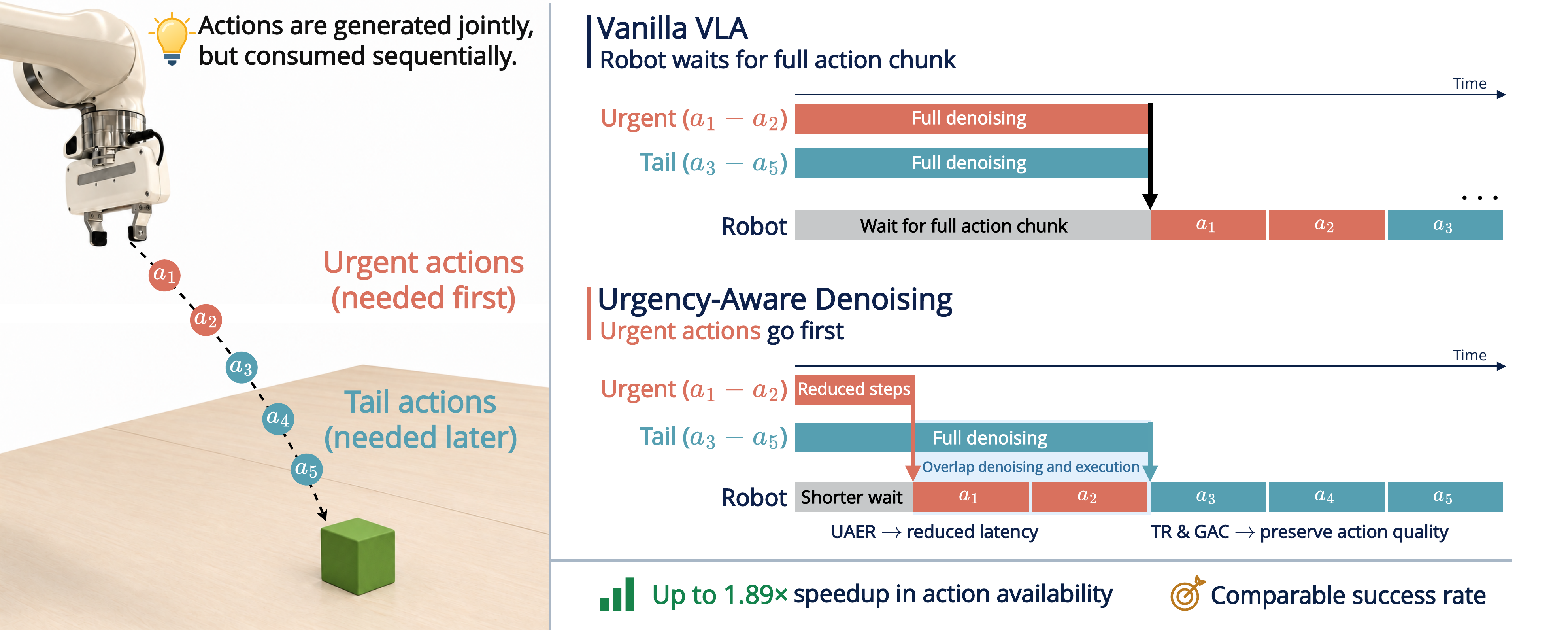}
    \caption{
    \textbf{Illustration of Urgency-Aware Denoising (UAD).}
    \textbf{Insight:} Actions are generated jointly but consumed sequentially, creating heterogeneous execution urgencies.
    \textbf{Method:} UAD releases urgent actions after fewer denoising steps and overlaps the remaining denoising with physical execution.
    \textbf{Modules:} Urgency-Aware Early Release (UAER) releases urgent actions early to reduce latency, Trajectory Reconciliation (TR) restores consistent joint denoising, and Ghost Action Correction (GAC) compensates for early-release errors using ghost actions.
    \textbf{Benefits:} UAD reduces action availability latency while maintaining comparable task success rates.
    }
    \label{fig:hook}
\end{figure}

This asymmetry suggests a new design principle: \textit{denoising computation should be allocated according to execution urgency rather than distributed uniformly across the chunk.} The first actions are the most urgent, as their generation latency directly determines when the robot can initiate physical execution. In contrast, subsequent tail actions have a larger temporal buffer before execution, allowing their refinement to naturally overlap with the physical execution of early actions. In practice, this principle concretely materializes as processing urgent actions with fewer denoising steps to release them early, while allowing tail actions to retain full denoising in the background.

However, realizing such heterogeneous denoising is non-trivial, as it introduces two core challenges. First, releasing urgent actions early inevitably introduces \textit{early-release errors}, causing action deviations from the baseline. Second, since actions within a chunk are coupled through shared attention in the DiT, altering the state trajectories of urgent actions disrupts the joint attention context. This induces severe \textit{trajectory inconsistency} and errors in the tail actions, even if those tail actions undergo full denoising steps. Consequently, naive early release fails to maintain action fidelity.

To address these challenges, we propose \textbf{Urgency-Aware Denoising (UAD)}, an inference-time framework that accelerates action availability while safeguarding action trajectory. Fig. \ref{fig:hook} illustrates the insight, method, and benefits of UAD. The framework comprises three core components:
(1) \textbf{Urgency-Aware Early Release (UAER)} performs fewer denoising steps for urgent actions to make them available immediately, allowing tail action denoising to overlap with physical execution.
(2) \textbf{Trajectory Reconciliation (TR)} reconstructs unified hidden state trajectories for urgent actions using already-computed velocities, restoring joint denoising consistency for tail actions with zero additional model evaluations.
(3) \textbf{Ghost Action Correction (GAC)} uses the non-executed continuation (``ghost action'') of released urgent actions to estimate early-release errors, dynamically compensating for them across remaining executable actions.
Together, these components effectively mitigate the trade-off between action latency and trajectory coherence, enabling rapid action availability while mitigating errors caused by early release.

In summary, our contributions are:
\begin{itemize}
\item \textbf{Urgency-Aware Inference Paradigm.}
We introduce a new inference-time perspective for real-time VLA control, which allocates denoising computation according to temporal execution urgency rather than uniformly across an action chunk. Driven by the observation that chunked actions are generated jointly but consumed sequentially, this principle enables urgent actions to be released significantly earlier, cutting action availability latency.
\item \textbf{Practical System Design.}
We make heterogeneous denoising practical through the proposed UAD framework. Beyond Urgency-Aware Early Release (UAER) for latency reduction, UAD resolves the resulting early-release errors and trajectory inconsistency via two core components: Trajectory Reconciliation (TR), which restores joint state alignment with zero additional DiT evaluations, and Ghost Action Correction (GAC), which dynamically compensates for execution deviations.
\item \textbf{Comprehensive Empirical Evaluation.}
We evaluate UAD across diverse VLA architectures, simulation benchmarks, and real-world robotic manipulation tasks. In simulation, UAD achieves up to a $1.89\times$ speedup in average action availability latency by overlapping background denoising with physical execution, while preserving task success rates comparable to vanilla inference with optimal denoising step configuration.
\end{itemize}

\section{Related Work}

\paragraph{DiT Acceleration for VLA Policies.}

A broad range of acceleration techniques has been explored for diffusion and flow-matching VLA policies. One line of work reduces the per-step evaluation cost via model quantization \citep{xu2026qvla,zhang2026quantvla}, layer skipping \citep{yang2026dysl}, pruning \citep{ji2026sparse, DBLP:conf/iccv/WuWCPX25}, distillation \citep{DBLP:conf/rss/PrasadLWZB24,DBLP:conf/icml/WangLMXFNFZBZ0025}, and feature caching \citep{yang2026efficientvla,ji2026block,DBLP:conf/iclr/ReussP0L25}. These architectural techniques are orthogonal to our denoising step allocation. Beyond model-level acceleration, a more closely related line of work leverages the domain-specific physical and temporal properties of robotic tasks to adaptively orchestrate inference execution \citep{yu2025d3p,salla2026learning,duan2025real,chun2026dynamic,DBLP:conf/icml/ChenLM0M0CZW0025}. For instance, D3P \citep{yu2025d3p} dynamically assigns denoising steps based on task-level action criticality, while POGP \citep{salla2026learning} employs adaptive early stopping based on predicted denoising gain. Concurrently, streaming architectures such as RTC \citep{black2026real} and asynchronous execution frameworks \citep{kim2026time,wang2026real} hide latency by overlapping action generation with physical execution. In contrast to these approaches, UAD operates along a novel dimension: it exploits temporal execution urgency to allocate heterogeneous denoising budgets within a jointly generated action chunk, reducing action availability latency while systematically mitigating the resulting trajectory errors.

\paragraph{Heterogeneous Action Chunk Generation.}
Recent works have explored heterogeneous generation across different positions within an action chunk to improve temporal modeling, action quality, or real-time reactivity. Diffusion Forcing \citep{chen2024diffusion} introduced per-position diffusion states by allowing different tokens to follow distinct noise levels, while subsequent methods extended this idea to robotic control through progressively varying refinement levels across the action horizon \citep{hoeg2025fast,chen2025responsive,jiang2025streaming,zhou2026latent}. More recent approaches further exploit position-dependent noise schedules for reactive control and cross-horizon action updating \citep{park2026pi,kim2026diffusion}. 
While these approaches share with UAD the high-level idea of introducing intra-chunk heterogeneity, they differ fundamentally in purpose and mechanism. Existing methods manipulate per-position noise levels or flow trajectories primarily for causal modeling and horizon refinement. In contrast, UAD allocates heterogeneous denoising computation explicitly according to execution urgency, and introduces TR and GAC to mitigate the resulting trajectory inconsistencies and early-release errors.

\section{Background}\label{sec_background}

In this section, we introduce the generation and execution lifecycle of flow-matching VLA policies, and analyze the fundamental trade-off between denoising fidelity and inference latency.

Under a receding-horizon control paradigm, a flow-matching VLA policy repeatedly generates control trajectories conditioned on multi-modal context. At each decision step, the policy ingests the current observation $o$ (comprising visual frames and sensory feedback) alongside a natural language instruction $\mathcal{L}$, and generates an action chunk $\mathbf{a} = [a^0, a^1, \ldots, a^{H-1}]$ containing $H$ future control steps. The robot executes only the leading prefix of $S$ actions ($S \le H$), captures a fresh observation, and re-invokes the policy. Because these $S$ executable actions are dispatched to physical hardware sequentially, earlier actions have greater execution urgency than later ones. This temporal asymmetry establishes distinct execution urgencies across actions within the same chunk.

Generating an action chunk corresponds to one DiT inference, consisting of $N$ sequential denoising steps. The policy initializes hidden state $\mathbf{x}_0 \sim \mathcal{N}(0, \mathbf{I})$, where $\mathbf{x}_k = [x_k^0, x_k^1, \ldots, x_k^{H-1}]$ denotes the hidden state at denoising step $k$, and $k=1,\ldots,N$. At step $k$, the DiT backbone predicts a velocity field $\mathbf{v}_k$ conditioned on the current state and context\footnote{We omit the flow-time argument $t$ for simplicity; see Appendix \ref{app:implementationsub1} for details.}:
\begin{equation}
\textbf{v}_k\leftarrow F_\theta(\mathbf{x}_{k-1},o,\mathcal{L}),
\end{equation}
where $\textbf{v}_k=[v_k^0,v_k^1,\ldots,v_k^{H-1}]$, and $v_k^i$ denotes the predicted velocity for the $i$-th action. Under a uniform Euler discretization scheme, each action state is integrated as:
\begin{equation}
x_{k}^i=
x_{k-1}^i
+
\frac{1}{N}  v_k^i,
\qquad
i=0,\ldots,H-1.
\end{equation}
After $N$ updates, the terminal state $\mathbf{x}_N = [x_N^0, \ldots, x_N^{H-1}]$ is released as the action chunk $\mathbf{a}$.

Increasing $N$ does not monotonically yield better performance. Across diverse VLA architectures and benchmarks, task success rates exhibit non-monotonic behavior with $N$ (see Appendix~\ref{app:denoising_budget}). An empirically optimal budget $N^* \in [2, 7]$ can be identified as the minimal step achieving peak success rates in settings sensitive to $N$. Consequently, a reasonable deployment strategy fixes $N = N^*$ for all generated actions.

However, action generation latency scales virtually linearly with $N$, augmented by a fixed overhead from vision-language encoding (Appendix~\ref{app:denoising_budget}). While aggressively truncating $N$ from $N^*$ down to a single step ($N=1$) promises a $2$--$7\times$ speedup in DiT inference, uniform step reduction degrades success rates by $3.9$-$10.7$ pp on sensitive tasks. This tension reveals a critical mismatch: uniform denoising treats all actions equally, ignoring their heterogeneous temporal urgencies. Since physical execution only requires immediate access to the earliest actions, it is unnecessary to reduce denoising steps uniformly across the entire action chunk. 
Instead, we can selectively truncate the denoising steps for urgent actions to enable immediate release, while preserving optimal $N^*$ for tail actions, accompanied by dedicated mechanisms to reconcile the resulting trajectory errors.

\section{Method}\label{sec:method}

In this section, we present UAD, an inference-time framework that exploits heterogeneous execution urgency within an action chunk to reduce action availability latency while preserving action quality.

\begin{figure}
    \centering
    \includegraphics[width=1.0\linewidth]{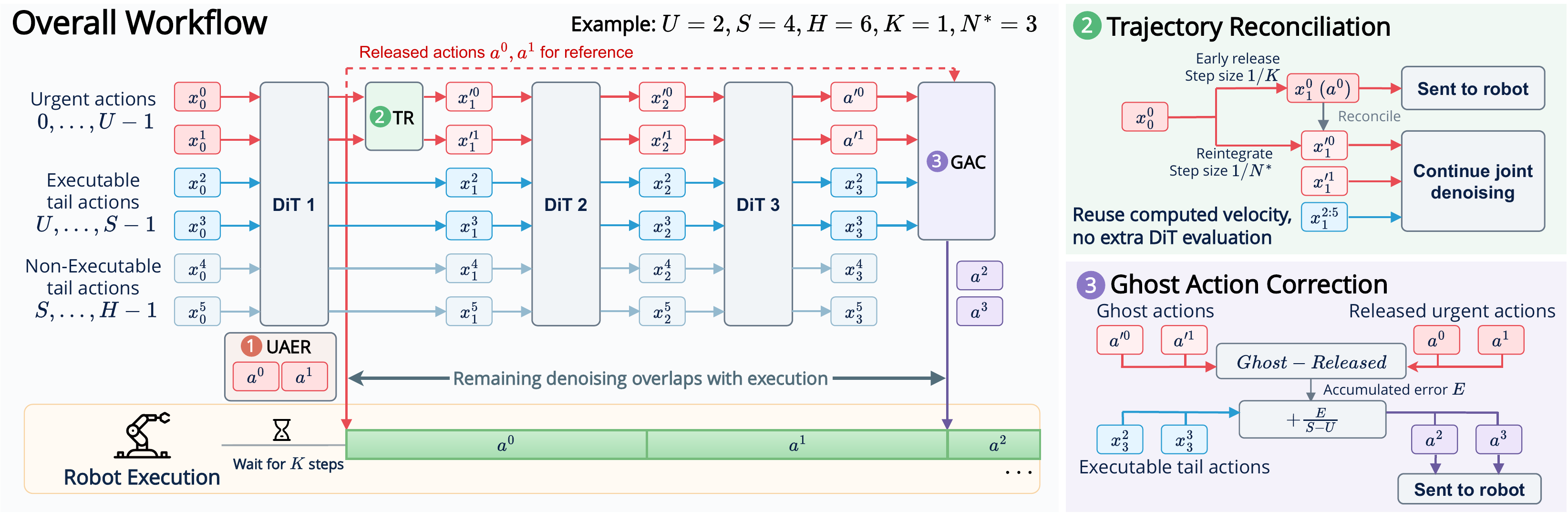}
    \caption{
    Overview of UAD
    }
    \label{fig:overview}
\end{figure}

\subsection{Overview of Urgency-Aware Denoising}

Fig.~\ref{fig:overview} illustrates the overall workflow of UAD.
An action chunk generated by a VLA policy can be categorized into three functionally distinct segments based on execution urgency: \textbf{urgent actions} needed immediately for physical execution, \textbf{executable tail actions} that have some temporal buffer, and \textbf{non-executable tail actions} (indexed by $S\leq i < H$, if $S<H$) that serve solely as temporal context.
UAD exploits this temporal urgency hierarchy to accelerate action availability while maintaining full trajectory quality.

Heterogeneous denoising, however, introduces two critical challenges.
First, releasing urgent actions early after fewer denoising steps introduces \textit{early-release errors} due to incomplete denoising.
Second, as action tokens interact via shared attention within the DiT, early release disrupts the alignment of denoising procedure between urgent and tail actions, causing \textit{trajectory inconsistency} that degrades subsequent tail action denoising.

UAD resolves these challenges through three integrated components, as structured in Fig.~\ref{fig:overview}.
\textbf{Urgency-Aware Early Release (UAER)} assigns a reduced denoising budget to urgent actions while retaining the full budget for tail actions, releasing urgent actions early to initiate physical execution and create an overlap window.
\textbf{Trajectory Reconciliation (TR)} restores consistent joint state evolution prior to subsequent DiT evaluations. By re-integrating already computed velocities, TR reconstructs compatible internal states for urgent actions with zero additional DiT evaluations, mitigating or even eliminating \textit{trajectory inconsistency}.
\textbf{Ghost Action Correction (GAC)} compensates for accumulated \textit{early-release errors}. As reconstructed urgent states continue joint denoising alongside the tail, they yield fully denoised \textbf{ghost actions}. GAC uses the deviation between ghost and released urgent actions to dynamically adjust the remaining executable tail actions.
Together, these components allow UAD to substantially reduce action availability latency while preserving the quality and execution accuracy of the generated trajectory.

\subsection{Urgency-Aware Early Release}

Each inference generates an action chunk of length $H$, of which only the first $S$ actions are executed before the next inference cycle is triggered. We designate the first $U$ actions ($U < S \le H$) as \textbf{urgent actions} and the remaining actions as \textbf{tail actions}, defining their index sets as
\begin{equation}
\mathcal{U}=\{0,\ldots,U-1\}, \qquad
\mathcal{T}=\{U,\ldots,H-1\}.
\end{equation}
The urgent actions are released early before full denoising completes, enabling their physical execution to overlap with the ongoing generation of the tail actions.

Under the vanilla policy, all actions undergo $N^*$ denoising steps, where $N^*$ is the empirically determined optimal setting. UAD instead assigns a reduced budget $K < N^*$ to urgent actions while retaining $N^*$ steps for tail actions:
\begin{equation}
N_i=
\begin{cases}
K, & i\in\mathcal{U},\\
N^*, & i\in\mathcal{T}.
\end{cases}
\end{equation}
Compared with the uniform Euler update introduced in Section~\ref{sec_background}, UAER preserves the same integration interval while assigning different numbers of denoising steps to urgent and tail actions, resulting in action-specific Euler step sizes. Specifically, the state of action $i$ evolves as
\begin{equation}
x_{k}^i=x_{k-1}^i+\frac{1}{N_i}v_k^i,\qquad k=1,\ldots,N_i.
\end{equation}
Upon completing $K$ denoising steps, the state $x_K^i$ of each urgent action $i \in \mathcal{U}$ is released directly as the executable action $a^i = x_K^i$. Meanwhile, tail actions continue their denoising updates until all $N^*$ steps are completed.

\paragraph{Latency Benefit.}
UAER releases urgent actions after only $K$ denoising steps, allowing their physical execution to overlap with the remaining denoising for tail actions. When the execution duration of $U$ urgent actions is sufficient to cover this remaining computation, the tail denoising latency is entirely masked. A formal latency model and detailed analysis are provided in Appendix~\ref{app:latency_analysis}.

\subsection{Trajectory Reconciliation}

\begin{wrapfigure}[14]{r}{0.45\linewidth}
    \centering
    \includegraphics[width=\linewidth]{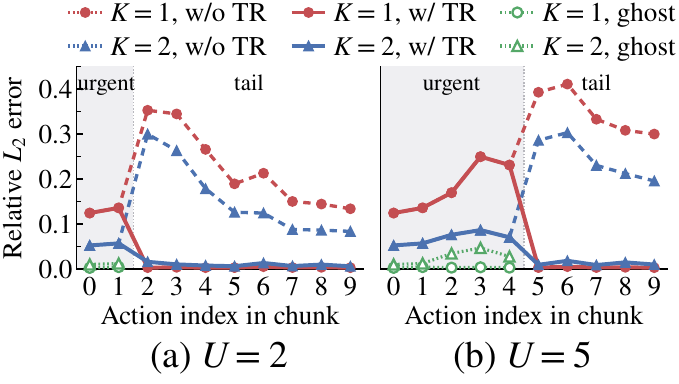}
    \caption{
    Relative $L_2$ error to vanilla inference ($N^*=3$).
    TR reduces tail action error while ghost states remain close to vanilla. 
    }
    \label{fig:trajectory_reconciliation}
\end{wrapfigure}

While UAER successfully reduces action availability latency, naively applying heterogeneous denoising severely degrades the quality of the entire action chunk. As empirically verified in Fig.~\ref{fig:trajectory_reconciliation}, naive early release introduces substantial relative $L_2$ errors across both urgent and tail actions. 
First, urgent actions naturally deviate from their fully denoised counterparts due to reduced denoising steps. 
Second, because flow-matching VLA policies model the entire action chunk jointly via shared attention, operating urgent actions on a larger step size ($1/K$) disrupts joint state alignment, propagating divergence to tail actions and corrupting trajectory consistency. This highlights the vital need for a dedicated mechanism to reconcile internal trajectories before subsequent joint denoising.

To eliminate such misalignment, we introduce Trajectory Reconciliation (TR), which reconstructs compatible internal states for urgent actions prior to subsequent denoising. Since DiT predicts velocity fields $v_k$ and action states are updated via Euler integration, previously computed velocities can be re-integrated under the canonical step size $1/N^*$. Specifically, for each urgent action $i \in \mathcal{U}$, TR computes a reconciled state $x^{\prime i}_K$ after $K$ steps as
\begin{equation}
x_K^{\prime i}=x_0^i+\frac{1}{N^*}\sum_{k=1}^{K} v_k^i .
\end{equation}
Crucially, this reconstruction requires \textbf{zero additional DiT evaluations}, as all velocity vectors $v_k^i$ ($k\leq K$) are already computed and cached.

Following reconciliation, the complete state evolution across the action chunk is summarized as
\begin{equation}
\begin{cases}
x_{k}^i = x_{k-1}^i + \frac{1}{K}v_k^i, & i\in\mathcal{U},\ k\leq K,\\
x_{k}^{\prime i} = x_{k-1}^{\prime i} + \frac{1}{N^*}v_k^i, & i\in\mathcal{U},\ k> K,\\
x_{k}^i = x_{k-1}^i + \frac{1}{N^*}v_k^i, & i\in\mathcal{T}.
\end{cases}
\end{equation}
By feeding the reconciled states $x^{\prime i}_K$ into subsequent DiT steps ($k > K$), TR reduces the joint-state misalignment before continued denoising. As demonstrated in Fig.~\ref{fig:trajectory_reconciliation}, TR sharply reduces the tail action error across different $U$ settings.

\paragraph{Exact Recovery for $K=1$.}
When $K=1$, since all actions share identical initial states and time conditions, the first DiT evaluation is identical to vanilla inference. Re-integrating this velocity with step size $1/N^*$ therefore exactly recovers the corresponding vanilla state, after which subsequent internal states follow the vanilla trajectory. Consequently, for $K=1$, tail actions incur theoretically \textit{zero additional error} relative to vanilla inference before applying GAC (Fig.~\ref{fig:trajectory_reconciliation}). For $K>1$, this exactness no longer holds because the cached velocities are evaluated along the heterogeneous early-release trajectory, and TR remains effective in substantially reducing tail action error empirically.

\paragraph{Ghost States and Ghost Actions.}
As released urgent actions are already dispatched for execution, their reconciled internal continuations $x^{\prime i}_k$ ($k \ge K$) are not executed directly. We define these continued trajectories as \textbf{ghost states}, and denote their fully denoised final states as \textbf{ghost actions}: $a^{\prime i} \triangleq x_{N^*}^{\prime i}$ ($i \in \mathcal{U}$). Ghost states maintain necessary temporal context for tail actions during joint denoising, while providing a reference that we exploit next for mitigating early-release errors.

\subsection{Ghost Action Correction}

While TR addresses the trajectory inconsistency introduced to tail actions, the early-release errors inherent to urgent actions remain uncompensated. Since released urgent actions have already been dispatched for physical execution, they cannot be corrected retroactively. To mitigate this residual execution drift, we introduce Ghost Action Correction (GAC), which estimates the accumulated early-release error via ghost actions and dynamically compensates for it through subsequent executable tail actions.

\paragraph{Error Estimation via Ghost Actions.}
Correcting early-release error requires an accurate reference of the ideal, fully denoised trajectory. As verified in Fig.~\ref{fig:trajectory_reconciliation}, Trajectory Reconciliation preserves joint temporal context, enabling ghost actions $a^{\prime i}$ to closely track the vanilla references with remarkably low relative $L_2$ error (achieving exact zero error when $K=1$). Consequently, $a'^i$ serves as an ideal proxy for the uncorrupted actions. Since physical execution errors accumulate over time, we quantify the total early-release error vector $\mathbf{E}$ across all $U$ urgent actions as
\begin{equation}
\mathbf{E} = \sum_{i=0}^{U-1} \left(a'^i - a^i\right) = \sum_{i=0}^{U-1} \left(x_{N^*}^{\prime i} - x_K^i\right),
\end{equation}
where $a^i = x_K^i$ is the released urgent action and $a^{\prime i} = x_{N^*}^{\prime i}$ is the corresponding ghost action.

\paragraph{Dynamic Compensation across Executable Tail Actions.}
Although executed urgent actions are immutable, VLA actions encode continuous spatial commands (\eg Cartesian translation and axis-angle rotation). In robotic control, small execution deviations can be compensated through subsequent corrective motion commands \citep{DBLP:conf/rss/ShiHZSPLLF24,wang2026real}. 
GAC leverages this physical compensability by uniformly distributing the accumulated error $\mathbf{E}$ over the remaining $S-U$ executable tail actions prior to their release. We adopt this simple allocation to preserve the total compensating displacement without introducing additional optimization or learned components:
\begin{equation}\label{eq:gac2}
a^i = x_{N^*}^{i} + \frac{\mathbf{E}}{S-U}, \qquad i = U, \ldots, S-1,
\end{equation}
where $x_{N^*}^{i}$ denotes the denoised state of tail action $i$ before correction.
This adjustment leaves executed urgent actions untouched while compensating for the accumulated command discrepancy over the control horizon.
In practice, this compensation is applied selectively based on physical semantics: additive dimensions (\eg Cartesian translation) follow Eq.~\ref{eq:gac2}, rotations are corrected through composition on $\mathrm{SO}(3)$, and decision-based dimensions (\eg gripper binary open/close) are masked to prevent unintended state toggling (detailed in Appendix \ref{app:implementationsub1}).
GAC is applicable when action dimensions admit compensation across consecutive control steps. For action representations defined relative to a fixed or chunk-level reference, we omit GAC and retain UAER with TR only.

In conclusion, TR and GAC together enable UAD to retain the latency benefit of UAER while addressing the two main sources of action degradation introduced by heterogeneous denoising.

\section{Evaluation}\label{sec:eval}

We evaluate UAD through simulation and real-world experiments, examining its effectiveness in reducing action availability latency while preserving task success rates. We implement UAD and baselines with the LeRobot framework \citep{cadene2026lerobot}. Implementation details for the benchmarks, UAD, and baselines, including parameter settings, are provided in Appendix~\ref{app:implementation}. Additional evaluation results are provided in Appendix~\ref{app:eval}.

\subsection{Experiment Setup}

\paragraph{Models and Benchmarks.}
We evaluate UAD with two representative VLA models across simulation and real-world robotic manipulation. In simulation, we evaluate SmolVLA \citep{shukor2025smolvla} on LIBERO-Plus \citep{fei2025libero} and Meta-World+ \citep{metaworldp}, and $\pi_{0.5}$ \citep{intelligence2025pi_} on Meta-World+. We further deploy $\pi_{0.5}$ on a real robotic platform.
The benchmark configuration follows our characterization of denoising step behavior in Appendix~\ref{app:denoising_budget}. For the simulation experiments, we use pretrained checkpoints finetuned for the corresponding environments from the LeRobot framework or the community \citep{ckpt_smolvla_libero_plus,ckpt_smolvla_metaworld,ckpt_pi05_metaworld}.

\paragraph{Baselines.}
We compare UAD with five baselines.
\textbf{Vanilla (\bm{$N^*$})} uses the empirically optimal number of denoising steps for each setting, providing the reference for task success rate and latency.
\textbf{Vanilla (\bm{$N=1$})} uniformly reduces denoising to a single step, serving as a direct comparison with aggressive step reduction.
We further compare UAD with three state-of-the-art acceleration methods: \textbf{D3P} \citep{yu2025d3p}, which dynamically selects denoising budgets; \textbf{RTC} \citep{black2026real}, which overlaps action generation with execution; and \textbf{BAC} \citep{ji2026block}, which adaptively reuses intermediate features at the block level across denoising steps to reduce redundant computation.

\paragraph{Metrics.}
We evaluate task performance using \textbf{success rate (SR)}, defined as the fraction of evaluation episodes completed successfully, and inference latency using \textbf{average action availability latency (AAL)}, which measures the waiting time from requesting an action until it becomes available, averaged over all executed actions.
AAL is therefore an action-level measure of exposed inference latency: given a fixed action duration $T_{\mathrm{act}}=1/f_{\mathrm{ctrl}}$, it directly corresponds to an idle-time fraction of $\mathrm{AAL}/(\mathrm{AAL}+T_{\mathrm{act}})$.
To facilitate comparison with the standard inference setting, we additionally report \textbf{$\Delta$SR}, the change in success rate relative to Vanilla ($N^*$) in percentage points, and \textbf{speedup}, computed as the ratio of the AAL of Vanilla ($N^*$) to that of each method.

\paragraph{Settings.}
Following the convention of existing research or the default LeRobot configuration, we use an action chunk size of $H=50$ and execute $S=10$ actions before replanning. For UAD, we set $K=1$ to minimize urgent-action latency while enabling lossless recovery of tail and ghost actions. We choose the smallest $U$ for which urgent-action execution fully overlaps the remaining tail denoising. All VLA inference and latency measurements are performed on an NVIDIA H20 GPU. Additional hardware and configuration details are provided in Appendix~\ref{app:implementation}.

\subsection{Results and Analysis}

\begin{table}[t]
\centering
\caption{
    Main results across three simulation settings.
    See \textbf{Metrics} for the definitions of metrics.
}
\label{tab:main_results}

\footnotesize
\setlength{\tabcolsep}{1.8pt}
\renewcommand{\arraystretch}{1.08}

\begin{tabular*}{\linewidth}{
    @{\extracolsep{\fill}}
    lcccccccccccc
    @{}
}

    \toprule

    &
    \multicolumn{4}{c}{\textbf{SmolVLA + LIBERO-Plus}}
    &
    \multicolumn{4}{c}{\textbf{SmolVLA + Meta-World+}}
    &
    \multicolumn{4}{c}{\textbf{$\bm{\pi_{0.5}}$ + Meta-World+}}
    \\
    \cmidrule(lr){2-5}
    \cmidrule(lr){6-9}
    \cmidrule(lr){10-13}

    Method
    & SR
    & AAL
    & $\Delta$SR
    & Speedup
    & SR
    & AAL
    & $\Delta$SR
    & Speedup
    & SR
    & AAL
    & $\Delta$SR
    & Speedup
    \\

    &
    (\%) $\uparrow$
    & (ms) $\downarrow$
    & (pp) $\uparrow$
    & ($\times$) $\uparrow$
    & (\%) $\uparrow$
    & (ms) $\downarrow$
    & (pp) $\uparrow$
    & ($\times$) $\uparrow$
    & (\%) $\uparrow$
    & (ms) $\downarrow$
    & (pp) $\uparrow$
    & ($\times$) $\uparrow$
    \\

    \midrule

    Vanilla ($N^*$)
    & 35.3 & 21.6 & 0.0 & 1.00
    & 70.8 & 14.4 & 0.0 & 1.00
    & 70.5 & 11.2 & 0.0 & 1.00
    \\

    Vanilla ($N=1$)
    & 27.3 & 11.2 & $-8.0$ & 1.93
    & 60.1 & 10.8 & $-10.7$ & 1.33
    & 66.6 & 9.0 & $-3.9$ & 1.24
    \\

    \midrule

    D3P
    & 36.3 & 17.6 & $+1.0$ & 1.23
    & 70.3 & 13.9 & $-0.5$ & 1.04
    & 70.5 & 11.6 & 0.0 & 0.97
    \\

    RTC
    & 8.8 & 2.7 & $-26.5$ & 8.00
    & 52.3 & 1.3 & $-18.5$ & 11.08
    & 59.8 & 1.5 & $-10.7$ & 7.47
    \\

    BAC
    & 35.3 & 17.3 & 0.0 & 1.25
    & 62.8 & 13.0 & $-8.0$ & 1.11
    & 58.5 & 8.4 & $-12.0$ & 1.33
    \\

    \midrule

    UAD (ours)
    & 36.5 & 11.4 & $+1.2$ & 1.89
    & 67.9 & 11.1 & $-2.9$ & 1.30
    & 69.2 & 8.5 & $-1.3$ & 1.32
    \\

    \bottomrule

\end{tabular*}

\end{table}

\begin{figure}[t]
    \centering
    \includegraphics[width=1\linewidth]{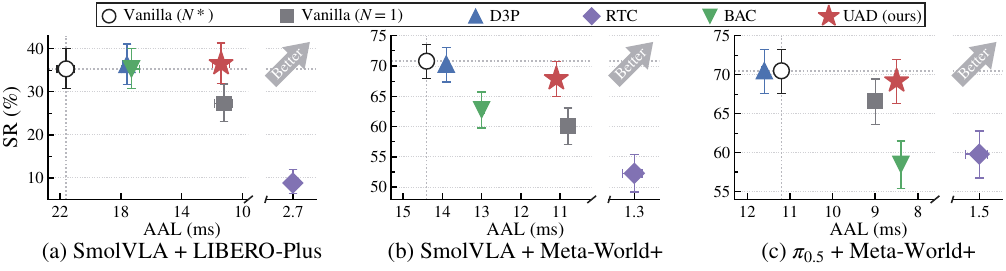}
    \caption{
        Success rate-latency trade-offs across three simulation settings.
        UAD achieves low AAL while retaining high SR.
        Broken horizontal axes accommodate RTC's substantially lower AAL. Error bars indicate Wilson 95\% CIs for SR, and episode bootstrap 95\% CIs for AAL. Some AAL CIs are too narrow to be visible at the plotted scale.
    }
    \label{fig:tradeoff}
\end{figure}

\paragraph{Overall Performance.}
As shown in Table~\ref{tab:main_results}, in three simulation settings, UAD achieves $1.30$--$1.89\times$ speedup in AAL over Vanilla ($N^*$), with SR changes ranging from $-2.9$ to $+1.2$ pp.
Compared with Vanilla ($N=1$), UAD improves SR by $9.2$, $7.8$, and $2.6$ pp, respectively, while achieving comparable AAL.
These results demonstrate that allocating denoising steps according to action urgency mitigates the performance loss caused by single-step inference.
D3P adopts conservative denoising step selection in the experiments, largely preserving SR but yielding only limited AAL speedups ($0.97$-$1.23\times$), whereas UAD achieves larger speedups across all three settings.
Compared with BAC, UAD consistently achieves higher SR with lower or comparable AAL.
RTC achieves very low AAL by fully overlapping VLA computation with action execution.
However, its SR drops substantially by $10.7$–$26.5$ pp relative to Vanilla ($N^*$) in the evaluated settings.
Overall, UAD offers a favorable balance between action availability latency and task success, as is shown in Fig. \ref{fig:tradeoff}.

\begin{figure}[t]
    \centering

    \begin{minipage}[t]{0.66\linewidth}
        \vspace{0pt}
        \centering

        \begin{minipage}[t]{0.485\linewidth}
            \centering
            \includegraphics[width=\linewidth]{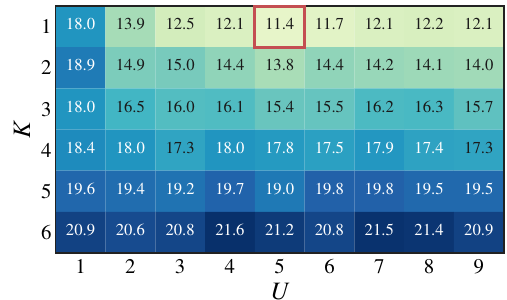}
            
            \small (a) AAL.
        \end{minipage}
        \hfill
        \begin{minipage}[t]{0.485\linewidth}
            \centering
            \includegraphics[width=\linewidth]{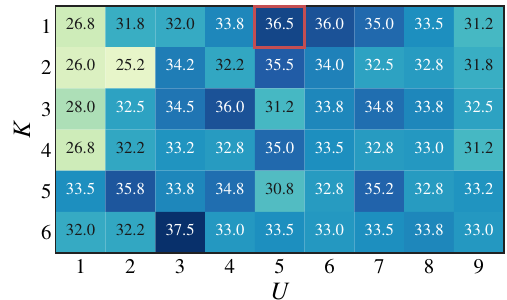}
            
            \small (b) SR.
        \end{minipage}

        \captionof{figure}{
            Parametric study of UAD under different urgent-action counts
            $U$ and urgent denoising depths $K$.
            The applied setting is marked by a red square.
        }
        \label{fig:parametric}
    \end{minipage}
    \hfill
    \begin{minipage}[t]{0.32\linewidth}
        \vspace{0pt}
        \centering

        \includegraphics[width=\linewidth]{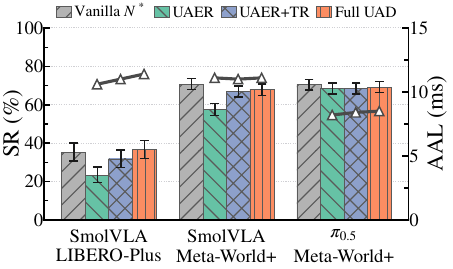}

        \captionof{figure}{
            Ablation study of the components in UAD.
            Bars: SR; markers: AAL; error bars: Wilson 95\% CI.
        }
        \label{fig:ablation}
    \end{minipage}

\end{figure}

\paragraph{Parameter Study.}
We study the two key parameters of UAD, $K$ and $U$, on SmolVLA with LIBERO-Plus, whose relatively large $N^*=7$ provides a broad design space. We evaluate all feasible settings ($K=1,\ldots,6$ and $U=1,\ldots,9$), with results shown in Fig. \ref{fig:parametric}.
Reducing $K$ lowers AAL by shortening the blocking time for urgent actions.
Increasing $U$ allows urgent action execution to hide more tail denoising, but provides little additional latency benefit once the remaining computation is fully overlapped.
For SR, very small or large $U$ tends to underperform intermediate values, particularly at small $K$.
Meanwhile, increasing $K$ yields no consistent SR improvement.
With TR and GAC, $K=1$ maintains consistently strong SR across a range of intermediate $U$ values while minimizing AAL, supporting our choice of $K=1$ with $U$ covering the remaining computation.

\begin{figure}[t]
    \centering

    \begin{minipage}[t]{0.55\linewidth}
        \vspace{0pt}
        \centering

        \includegraphics[
            width=\linewidth,
            height=0.1\textheight,
            keepaspectratio
        ]{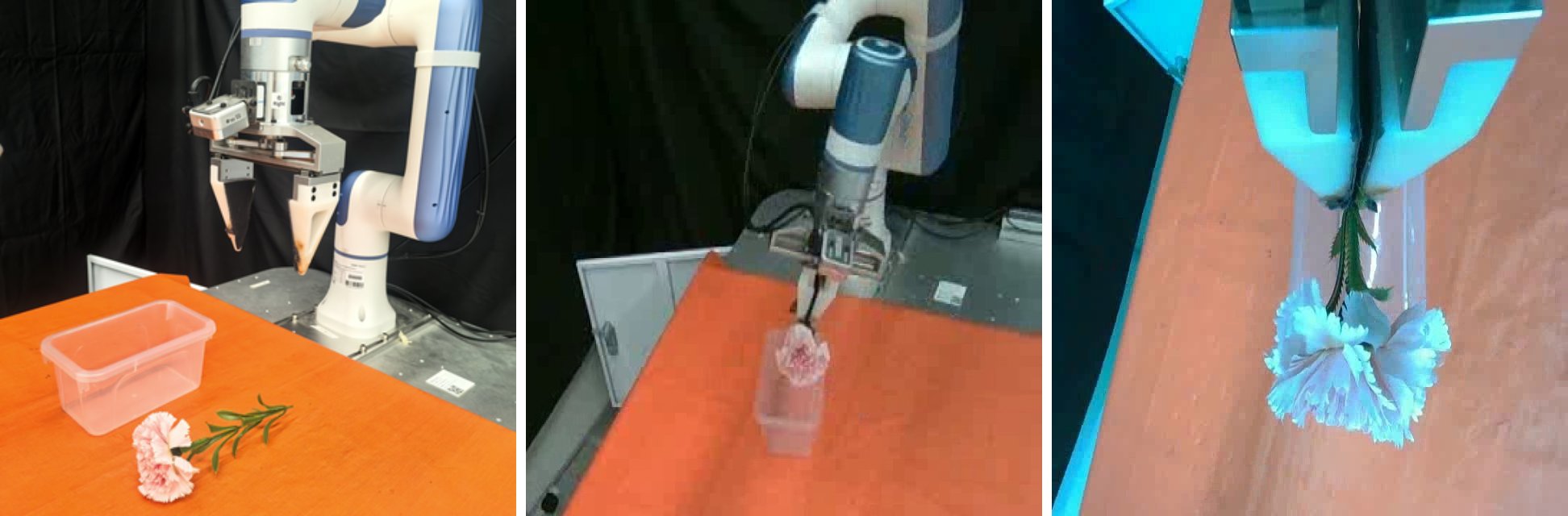}

        \caption{
            Real-world deployment on Dobot Nova2.
        }
        \label{fig:real_world}
    \end{minipage}
    \hfill
    \begin{minipage}[t]{0.4\linewidth}
        \vspace{0pt}
        \centering

        \captionof{table}{Real-world deployment results.}
        \label{tab:real_world}

        \small
        \setlength{\tabcolsep}{2pt}
        \renewcommand{\arraystretch}{1.1}

        \begin{tabular*}{\linewidth}{
            @{\extracolsep{\fill}}lcc@{}
        }
            \toprule
            Method & SR (\%)$\uparrow$ & AAL (ms)$\downarrow$ \\
            \midrule
            Vanilla ($N=10$) & 84 & 34.3 \\
            Vanilla ($N=1$)  & 71 & 14.9 \\
            \textbf{UAD}     & 84 & 15.2 \\
            \bottomrule
        \end{tabular*}
    \end{minipage}

\end{figure}

\paragraph{Ablation Study.}
UAD comprises urgency-aware early release (UAER), trajectory reconciliation (TR), and ghost action correction (GAC).
UAER accelerates inference but may degrade action quality. TR addresses tail-action errors, while GAC uses ghost actions obtained through TR to further preserve action quality.
We compare UAER alone, UAER+TR, and full UAD in Fig.~\ref{fig:ablation}.
In both SmolVLA settings, UAER reduces SR, TR partially recovers the loss, and GAC further restores SR to a level comparable to Vanilla ($N^*$).
For $\pi_{0.5}$ on Meta-World+, UAER causes little SR degradation, leaving limited room for TR and GAC to improve performance, consistent with their role in recovering behavior under $N^*$ denoising.
AAL remains similar across variants in all settings, indicating negligible additional exposed latency from TR and GAC.

\paragraph{Real-World Deployment.}
As shown in Fig. \ref{fig:real_world}, we deploy UAD on a Dobot Nova2 robot using finetuned $\pi_{0.5}$ checkpoints across five tasks. We compare against vanilla inference with $N=10$, the default deployment configuration, and $N=1$, an aggressive acceleration baseline. As shown in Table~\ref{tab:real_world}, UAD reduces AAL from 34.3 to 15.2~ms, closely matching the 14.9~ms achieved with $N=1$ and yielding a $2.26\times$ speedup. UAD also maintains the default baseline's 84\% success rate. In contrast, $N=1$ reduces success to 71\%. These results demonstrate that UAD translates its latency benefits to real-world control while preserving task performance.

\section{Conclusion}

We introduced \textbf{Urgency-Aware Denoising (UAD)}, an inference-time framework for accelerating flow-matching VLA policies by exploiting the heterogeneous execution urgency within an action chunk. Rather than uniformly allocating denoising computation to actions that are consumed at different times, UAD releases urgent actions early and overlaps the remaining denoising with physical execution, while Trajectory Reconciliation (TR) and Ghost Action Correction (GAC) preserve the quality of the resulting trajectory. Across multiple VLA models and robotic manipulation benchmarks, UAD consistently reduces action availability latency while maintaining success rates comparable to vanilla inference, achieving up to a $1.89\times$ speedup in simulation and similarly strong latency reductions in real-world deployment. More broadly, our results suggest that \emph{when} an action is needed provides an important dimension for allocating inference computation, offering a complementary direction for improving the real-time efficiency of VLA control.

\subsection*{AI use statement}

In this work, we used generative AI tools for designing or providing feedback on research methodology or experiments, implementing methods, interpreting results, and assisting with translation. We did not use generative AI tools for generating synthetic datasets, developing the core theoretical or conceptual framework of the work, formulating mathematical claims, or proposing or refining research hypotheses. Tasks related to providing critical ingredients for mathematical proofs, assisting with proof writing, cleaning or reformatting datasets, and qualitative or thematic data analysis were not applicable to this work.
Additionally, we used generative AI tools to edit the background of a photograph of the robotic platform for use in Fig. \ref{fig:hook}, create or edit software code, assist with drafting and editing parts of the manuscript, summarize and identify relevant literature, source information, and suggest improvements to the structure and readability of the paper.
We reviewed all AI-assisted work. All content ultimately included in the paper was decided and evaluated by the authors, and AI-generated content was used only as reference. All AI-generated code used in the implementation was verified by the authors for correctness. We take responsibility for the final content of this work, including text, claims or artifacts produced with the aid of generative AI.

\subsection*{Reproducibility statement}

We provide the information needed to reproduce our method and experimental results throughout the main paper and appendix. The formulation and inference procedure of UAD are described in Section~\ref{sec:method}, with detailed implementation procedures provided in Appendix~\ref{app:implementationsub1}. The implementation and adaptation of the baselines are described in Appendix~\ref{app:impl2}. The evaluation protocols, configurations, and testbed implementation details are provided in Appendices~\ref{app:impl3} and~\ref{app:impl4} for the simulated environments and real-world deployment, respectively. 
Anonymous source code and evaluation scripts of the simulated environments are provided in the supplementary materials.

\bibliography{iclr2027_conference}
\bibliographystyle{iclr2027_conference}

\appendix

\section{Effect of the Number of Denoising Steps}
\label{app:denoising_budget}

We evaluate SmolVLA and $\pi_{0.5}$ on LIBERO, LIBERO-Plus, and Meta-World+, yielding six model-benchmark settings.
For each setting, we use a checkpoint finetuned on the corresponding benchmark \citep{ckpt_smolvla_libero,ckpt_smolvla_libero_plus,ckpt_smolvla_metaworld,ckpt_pi05_libero,ckpt_pi05_metaworld}, and vary the number of uniform denoising steps $N$ from $1$ to $10$.
We measure both task success rate and action availability latency.
These experiments characterize how the denoising budget affects the performance-latency trade-off and guide the selection of model-benchmark settings and the optimal denoising steps $N^*$ used in our main evaluation.

\newcommand{\uadBudgetImage}[1]{%
    \IfFileExists{#1}{%
        \includegraphics[width=\linewidth]{#1}%
    }{%
        \includegraphics[width=\linewidth]{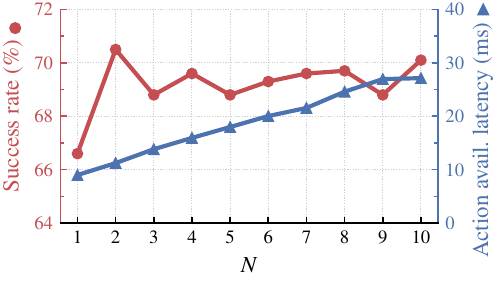}%
        \par\small\textit{Placeholder: duplicated image}%
    }%
}

\begin{figure}[t]
    \centering

    \begin{subfigure}[t]{0.30\linewidth}
        \centering
        \uadBudgetImage{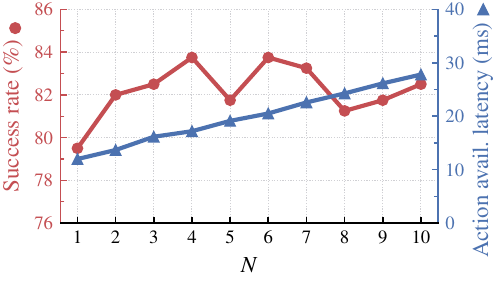}
        \caption{SmolVLA, LIBERO}
        \label{fig:budget_smolvla_l}
    \end{subfigure}
    \hfill
    \begin{subfigure}[t]{0.30\linewidth}
        \centering
        \uadBudgetImage{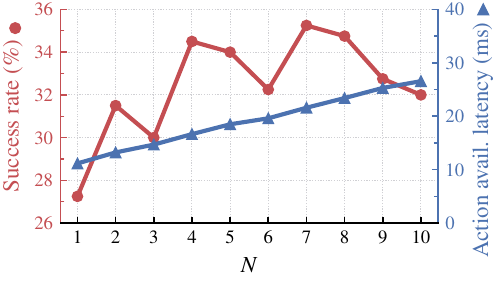}
        \caption{SmolVLA, LIBERO-Plus}
        \label{fig:budget_smolvla_lp}
    \end{subfigure}
    \hfill
    \begin{subfigure}[t]{0.30\linewidth}
        \centering
        \uadBudgetImage{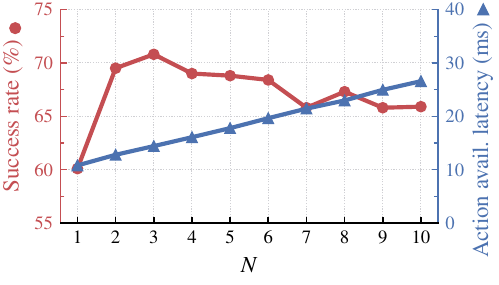}
        \caption{SmolVLA, Meta-World+}
        \label{fig:budget_smolvla_mw}
    \end{subfigure}

    \medskip

    \begin{subfigure}[t]{0.30\linewidth}
        \centering
        \uadBudgetImage{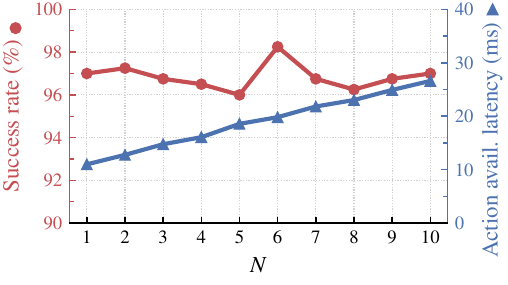}
        \caption{$\pi_{0.5}$, LIBERO}
        \label{fig:budget_pi05_l}
    \end{subfigure}
    \hfill
    \begin{subfigure}[t]{0.30\linewidth}
        \centering
        \uadBudgetImage{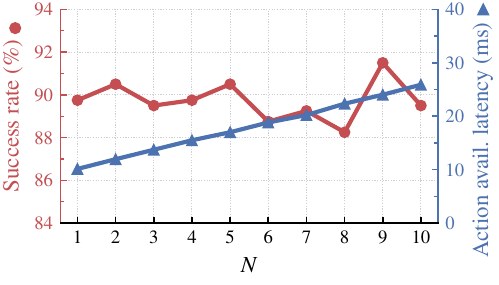}
        \caption{$\pi_{0.5}$, LIBERO-Plus}
        \label{fig:budget_pi05_lp}
    \end{subfigure}
    \hfill
    \begin{subfigure}[t]{0.30\linewidth}
        \centering
        \uadBudgetImage{fig/bac_pi05_mw.pdf}
        \caption{$\pi_{0.5}$, Meta-World+}
        \label{fig:budget_pi05_mw}
    \end{subfigure}

    \caption{
        Task success rate (red circles, left axis) and action availability
        latency (blue triangles, right axis) for different numbers of
        denoising steps $N$ across six model-benchmark settings.
    }
    \label{fig:background_budget_sweep}
\end{figure}

\paragraph{Task Performance and Benchmark Setting.}
Fig.~\ref{fig:background_budget_sweep} shows that the effect of the denoising budget strongly depends on the model-benchmark combination.
When a model is sufficiently strong for a relatively easy benchmark, task performance can become saturated and consequently insensitive to $N$.
For example, $\pi_{0.5}$ achieves consistently high success rates on LIBERO across the entire sweep, making changes in denoising quality difficult to distinguish from the task metric alone.
Such saturated settings provide limited information for studying the trade-off between denoising steps and action quality.
For each setting, we define $N^*$ as the smallest $N$ attaining the highest success rate and use it as the vanilla reference.
We therefore focus our main evaluation on settings that remain non-saturated and exhibit a clear sensitivity to the denoising steps:
SmolVLA on LIBERO-Plus ($N^*=7$) and Meta-World+ ($N^*=3$), and $\pi_{0.5}$ on Meta-World+ ($N^*=2$).
In these settings, success rate is non-monotonic in $N$, indicating that simply increasing the number of denoising steps does not necessarily improve task performance.
Reducing denoising steps uniformly from $N^*$ to $N=1$ decreases success rate by $3.9$-$10.7$ pp across these settings.
This establishes a regime where reducing denoising computation reduces latency but noticeably degrades action quality.
Results for the three additional settings not selected for the main evaluation are presented in Appendix \ref{app:sub_additional_settings}.

\paragraph{Latency.}
Across all six settings, average action availability latency increases approximately linearly with $N$, with a nonzero intercept.
This trend reflects the combination of repeated DiT evaluations and fixed inference overhead, including VLM context processing.
Consequently, reducing $N$ directly lowers denoising latency, but the fixed overhead limits the corresponding reduction in end-to-end action availability latency.
Together with the performance results above, this motivates reducing the computation exposed before execution rather than uniformly reducing the denoising steps of the entire action chunk.

\section{Latency Analysis of Urgency-Aware Early Release}
\label{app:latency_analysis}

We formulate a simplified timing model to explain how UAER reduces action availability latency by overlapping denoising computation with action execution.
Let $T_{\mathrm{VLM}}$ denote the latency of vision-language context encoding, $T_D$ the latency of a single DiT backbone evaluation, and $T_{\mathrm{act}}$ the execution duration of one action.
Each control cycle executes $S$ actions before invoking the policy again.
The analysis assumes constant computation and action execution times, ideal overlap between computation and execution, and no additional scheduling or synchronization overhead.
For simulation, $T_{\mathrm{act}}$ corresponds to the fixed overlap budget assigned to each action.

\paragraph{Vanilla Inference.}
Under vanilla inference, all $H$ actions undergo $N^*$ denoising steps and become available simultaneously.
The controller waits for VLM encoding and all $N^*$ DiT evaluations before execution begins.
Averaged over the $S$ executed actions, the action availability latency is:
\begin{equation}
T_{\mathrm{avail, vanilla}}
=
\frac{T_{\mathrm{VLM}} + N^*T_D}{S}.
\label{eq:latency_vanilla}
\end{equation}

\paragraph{Urgency-Aware Early Release.}
UAER releases the first $U$ urgent actions after $K < N^*$ denoising steps, allowing execution to begin after $T_{\mathrm{VLM}} + KT_D$.
Executing these urgent actions provides an overlap window of $UT_{\mathrm{act}}$ for the remaining $(N^*-K)$ DiT evaluations.
The controller incurs additional waiting only if tail denoising exceeds this window.
The resulting average action availability latency is:
\begin{equation}
T_{\mathrm{avail, UAER}}
=
\frac{
T_{\mathrm{VLM}} + KT_D
+ \max\left((N^*-K)T_D - UT_{\mathrm{act}},\,0\right)
}{S}.
\label{eq:latency_uaer}
\end{equation}
Thus, $K$ determines the initial waiting time, whereas $U$ determines the execution window available to overlap the remaining denoising.
When $UT_{\mathrm{act}} \ge (N^*-K)T_D$, tail denoising is fully hidden under this model, reducing the average action availability latency to $(T_{\mathrm{VLM}} + KT_D)/S$.

\paragraph{Practical Measurement and Parameter Selection.}
The timing model provides intuition for latency reduction, while actual latency depends on runtime overheads and timing variability.
We therefore base our empirical evaluation on measured AAL and blocking latency rather than the analytical estimates.
For each model-benchmark setting, we select the smallest $U$ whose measured average tail blocking time is below $0.2$\,ms, as detailed in Appendix~\ref{app:implementationsub1}.
This criterion keeps residual waiting negligible while limiting the number of actions released with fewer denoising steps.

\section{Implementation and Evaluation Details}\label{app:implementation}
\subsection{UAD Implementation}\label{app:implementationsub1}

We implement UAD within LeRobot for SmolVLA and $\pi_{0.5}$ by modifying their inference procedures, without introducing additional trainable parameters or requiring further training.

\paragraph{Per-Action Time Conditioning.}
Standard flow-matching inference conditions all action tokens on a shared flow time $t$. To support heterogeneous denoising schedules while retaining a single joint DiT forward pass, we extend the models to accept a separate time condition for each action token. For SmolVLA, the corresponding time embedding is provided to each action token before the existing action MLP; for $\pi_{0.5}$, it conditions the adaptive normalization layers of the action expert. These modifications reuse the pretrained weights and introduce no additional trainable parameters.

\paragraph{Asynchronous Execution.}
Once the urgent actions complete their assigned denoising steps, we synchronize the GPU and release them to the execution queue. Tail denoising then continues asynchronously on a separate CUDA stream while the controller executes the available actions. If tail computation is not completed before the urgent actions are consumed, execution waits for the remaining actions. After $S$ actions are executed, the controller replans from the latest observation.

\paragraph{TR and GAC.}
TR caches the initial noise and previously predicted velocities to reconstruct the internal urgent-action states without additional DiT evaluations, while keeping the released actions unchanged. GAC then uses the resulting ghost actions to correct the remaining executable actions, also without additional DiT evaluations. Corrections are applied in the appropriate action dimensions for each benchmark; in particular, rotational discrepancies are handled through composition on $\mathrm{SO}(3)$ rather than direct Euclidean subtraction.

\begin{wrapfigure}[12]{r}{0.35\linewidth}
    \centering
    \includegraphics[width=\linewidth]{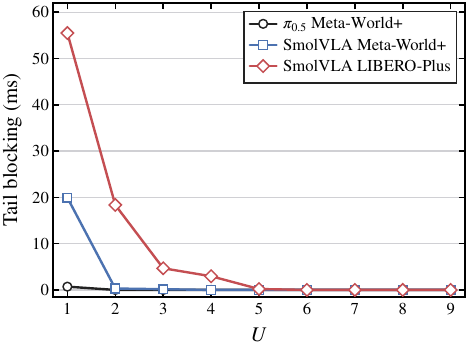}
    \caption{
        Average tail blocking time for different values of $U$ at $K=1$.
    }
    \label{fig:u_sweep}
\end{wrapfigure}
\paragraph{Parameter Configuration.}
Unless otherwise specified, we use $H=50$, $S=10$, and set the urgent denoising depth to $K=1$.
For each model-benchmark setting, we evaluate $U \in \{1,\ldots,S-1\}$ and measure the average blocking time of tail actions, as shown in Fig.~\ref{fig:u_sweep}.
We select the smallest $U$ that reduces this average waiting time below $0.2$\,ms, allowing the remaining denoising computation to overlap with urgent action execution with negligible residual blocking.
This yields $U=5$ for SmolVLA on LIBERO-Plus, $U=3$ for SmolVLA on Meta-World+, and $U=2$ for $\pi_{0.5}$ on Meta-World+.
Tail actions use the corresponding denoising budget $N^*$ selected in Appendix~\ref{app:denoising_budget}.

\subsection{Baseline Implementations}\label{app:impl2}

All baselines use the same pretrained VLA checkpoints and observation/action processing pipelines as UAD.
Unless otherwise specified, we use the same action horizon $H=50$ and execution interval $S=10$.
Baseline-specific adaptations and configurations are described below.

\paragraph{Vanilla.}
Vanilla ($N^*$) and Vanilla ($N=1$) use the standard LeRobot inference with different denoising steps.
Both generate the complete action chunk using Euler integration and execute the first $S$ actions before replanning.
The two variants use $N^*$ and one denoising step, respectively.

\paragraph{BAC.}
We implement BAC following its official open-source implementation \citep{github_bac}.
We retain its ACS and BUA mechanisms for adaptive recomputation and residual-cache updates, and adapt them to flow-matching VLAs by indexing the caching schedule over Euler steps and applying residual caching within the action expert.
Caches are reset for every action chunk.
We retain $N^*$ Euler steps and evaluate all ACS recomputation budgets $m \in \{1,\ldots,N^*-1\}$.
Among configurations that reduce AAL relative to vanilla inference, $m=\lceil N^*/2\rceil$ achieves the highest SR in all three simulation settings and is therefore used in our main comparison.
BUA is enabled for the five highest-drift blocks.
The caching schedule is calibrated offline on held-out rollouts using seeds distinct from evaluation, without additional policy training.

\paragraph{RTC.}
We use the original inference-time guided RTC implementation provided in LeRobot,
following its standard inference design and configuration. We retain its flow-matching
guidance and asynchronous action-queue update. Inference proceeds concurrently with
action execution; when a new chunk becomes available, actions that became stale during
inference are discarded before the remaining actions are inserted into the queue.

\paragraph{D3P.}
We reproduce D3P based on the original paper.
We adapt its learned denoising-step selection to pretrained flow-matching VLAs.
We let the VLA policy remain frozen, while a lightweight adaptor is trained with PPO to dynamically select the number of Euler intervals advanced at each denoising iteration.
Specifically, we divide the flow trajectory into $M=N^*$ uniform intervals.
At each denoising iteration, the adaptor selects an integer stride between one and the number of remaining intervals, and one velocity evaluation advances the action state across the selected range using the corresponding Euler step size.
We train a separate adaptor for each model-benchmark setting on a single NVIDIA H20 GPU for eight hours.
The reward combines task performance with a computation penalty based on the number of velocity evaluations, replacing the original diffusion-specific computation objective with its flow-matching counterpart.
At evaluation time, we use the final checkpoint and include adaptor inference overhead in the reported AAL.

\subsection{Simulation Evaluation Protocol}\label{app:impl3}

\paragraph{Benchmarks and Episodes.}
For LIBERO-Plus, we evaluate a fixed subset of 400 variants, with 100 sampled from each of the Spatial, Object, Goal, and Long suites, evenly covering all seven variation categories. We run one episode per variant using its designated initial state and the standard LeRobot evaluation horizon. Each episode begins with 10 settling steps that are excluded from evaluation. For Meta-World+, we evaluate 50 tasks with 20 episodes per task, yielding 1,000 episodes in total. Initial conditions vary across episodes. All methods use identical evaluation tasks and episode configurations, with a base random seed of 42. Episodes terminate upon environment-defined success or when the maximum horizon is reached.

\paragraph{Control-Time Alignment.}
The wall-clock duration of a simulation step can differ substantially from the physical execution duration of an action. Without explicit alignment, slow simulation or rendering could artificially enlarge the computation window available to asynchronous methods. We therefore assign each executed action a fixed overlap budget of $T_{\mathrm{act}}=1/f_{\mathrm{ctrl}}$, corresponding to 50\,ms for LIBERO-Plus at 20\,Hz and 12.5\,ms for Meta-World+ at 80\,Hz. For methods that overlap inference with execution, including UAD and RTC, background computation is limited by this prescribed budget rather than by simulator wall-clock time. Consequently, each action provides exactly one $T_{\mathrm{act}}$ of potential overlap, independent of how quickly or slowly the simulator executes the corresponding step.

\paragraph{Latency Measurement.}
We measure AAL as the delay from requesting an action until it becomes available to the controller, averaged over executed actions. The measurement includes exposed policy computation and waiting for unavailable actions, while excluding computation hidden behind action execution. Model loading, environment stepping, and rendering are excluded, and GPU synchronization is used to measure completed computation. 
All VLA inference is performed on NVIDIA H20 GPUs, with each inference process running on a single GPU. For latency measurement, we use one evaluation process on one exclusive H20 GPU with an inference batch size of one.
We evaluate all 50 Meta-World+ tasks and eight representative LIBERO-Plus variants spanning the four suites for latency measurement. 
We report steady-state AAL, excluding the first episode of each evaluation process as warm-up for all methods. This exclusion applies only to latency, and success rates are computed using all evaluation episodes.
We compute 95\% confidence intervals by resampling episodes with replacement (8,000 replicates) and taking the 2.5th and 97.5th percentiles of the bootstrap distribution.

\subsection{Real-World Deployment}\label{app:impl4}

\begin{figure}[t]
    \centering
    \includegraphics[width=\linewidth]{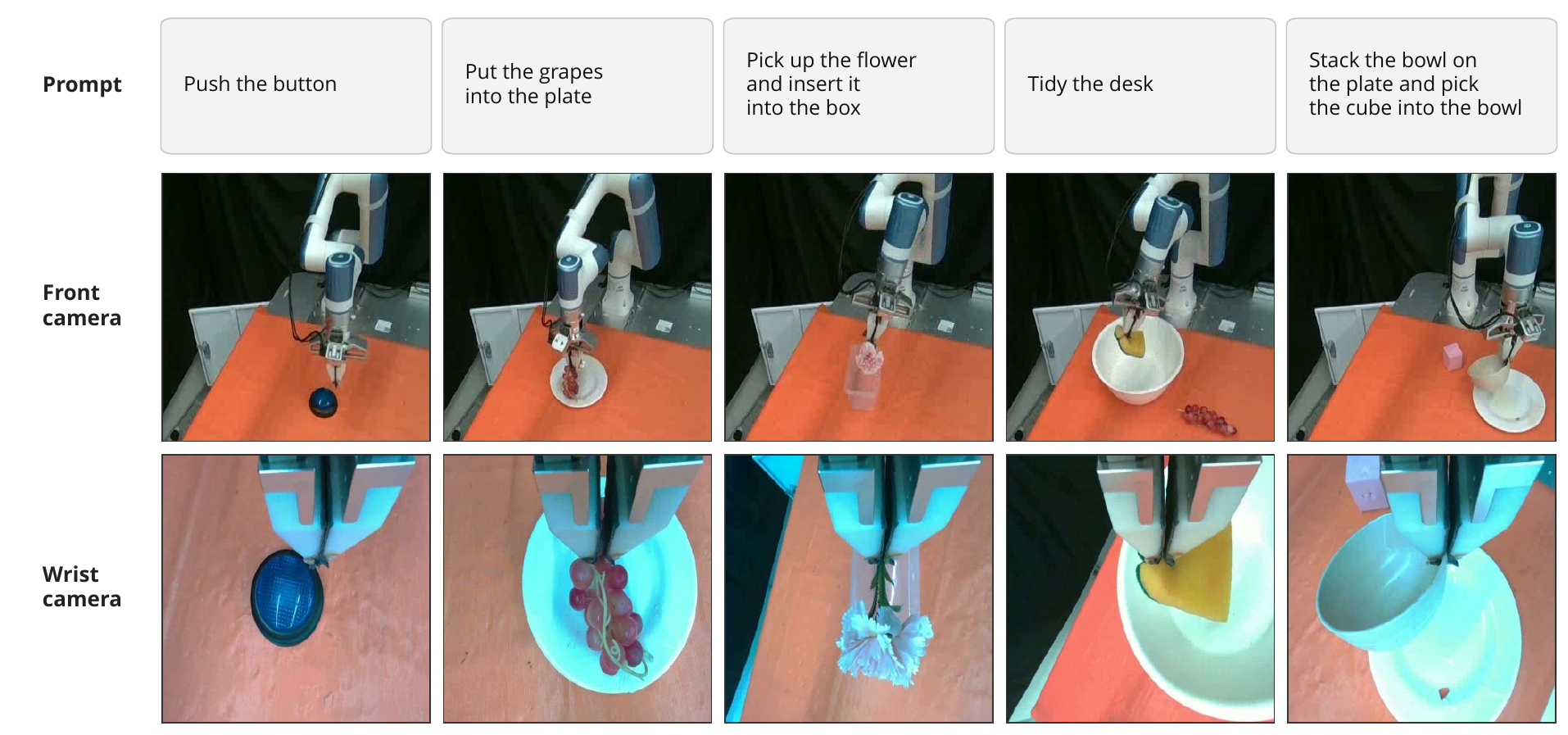}
    \caption{
        Real-world evaluation tasks and camera observations.
        We evaluate five manipulation tasks using front lateral and wrist-mounted camera views. Camera views are cropped.
    }
    \label{fig:real_task_gallery}
\end{figure}

\paragraph{Hardware Setup.}
We conduct real-world experiments on a Dobot Nova 2 robotic manipulation platform equipped with a front lateral camera (D455) and a wrist-mounted camera (D405). The two cameras use the existing calibrated configuration of the shared robotic platform, with the front-lateral view providing global workspace observations and the wrist view capturing close-range information around the end effector. VLA inference is performed on a single NVIDIA H20 GPU. The policy takes both camera views together with robot proprioception as input. All compared methods share the same robot platform, observations, action interface, and pretrained $\pi_{0.5}$ policy.

\paragraph{Tasks.}
We evaluate five tabletop manipulation tasks covering diverse interaction patterns, as illustrated in Fig.~\ref{fig:real_task_gallery}. Each task is specified by a natural-language instruction and requires the policy to coordinate the front-lateral and wrist-camera observations. We run 20 episodes per task, for a total of 100 real-world evaluation episodes. Together, the tasks cover contact interaction, pick-and-place, object rearrangement, and multi-stage manipulation.

\paragraph{Deployment and Evaluation.}
We deploy UAD on the pretrained $\pi_{0.5}$ policy. 
In this real-world setup, each action is represented as a relative change with respect to the beginning of the action chunk, rather than incrementally with respect to the preceding action. The additive compensation in GAC therefore does not apply to this action representation. Accordingly, our UAD deployment uses UAER together with TR, without GAC.
We compare UAD against Vanilla ($N=10$) and Vanilla ($N=1$) under identical hardware and control settings. We measure success rate manually, and AAL from the delay between an action request and the corresponding action becoming available for execution. Unlike in simulation, no control-time alignment is required: denoising computation overlaps directly with the physical execution of actions, and the resulting overlap window is determined by the robot's actual execution time.

\section{Extra Evaluation Results}\label{app:eval}

In this part, we present extra evaluation results not presented in Section \ref{sec:eval}.

\begin{table}[t]
\centering
\caption{
    Action availability latency (AAL) and DiT blocking latency (DBL)
    for the evaluated settings and methods.
}
\label{tab:aal_dbl}

\footnotesize
\setlength{\tabcolsep}{2.2pt}
\renewcommand{\arraystretch}{1.08}

\begin{tabular*}{\linewidth}{
    @{\extracolsep{\fill}}
    lcccccccc
    @{}
}

    \toprule

    &
    \multicolumn{2}{c}{\textbf{SmolVLA}}
    &
    \multicolumn{2}{c}{\textbf{SmolVLA}}
    &
    \multicolumn{2}{c}{\textbf{$\bm{\pi_{0.5}}$}}
    &
    \multicolumn{2}{c}{\textbf{Real-World}}
    \\
    &
    \multicolumn{2}{c}{\textbf{LIBERO-Plus}}
    &
    \multicolumn{2}{c}{\textbf{Meta-World+}}
    &
    \multicolumn{2}{c}{\textbf{Meta-World+}}
    &
    \multicolumn{2}{c}{\textbf{Deployment}}
    \\
    \cmidrule(lr){2-3}
    \cmidrule(lr){4-5}
    \cmidrule(lr){6-7}
    \cmidrule(lr){8-9}

    Method
    & AAL
    & DBL
    & AAL
    & DBL
    & AAL
    & DBL
    & AAL
    & DBL
    \\

    &
    (ms) $\downarrow$
    & (ms) $\downarrow$
    & (ms) $\downarrow$
    & (ms) $\downarrow$
    & (ms) $\downarrow$
    & (ms) $\downarrow$
    & (ms) $\downarrow$
    & (ms) $\downarrow$
    \\

    \midrule

    Vanilla ($N^*$)
    & 21.6 & 120.3
    & 14.4 & 52.4
    & 11.2 & 47.1
    & 34.3 & 216.6
    \\

    Vanilla ($N=1$)
    & 11.2 & 18.2
    & 10.8 & 17.9
    & 9.0 & 24.6
    & 14.9 & 19.9
    \\

    \midrule

    D3P
    & 17.6 & 122.2
    & 13.9 & 53.7
    & 11.6 & 36.1
    & - & -
    \\

    RTC
    & 2.7 & 9.6
    & 1.3 & 5.2
    & 1.5 & 7.0
    & - & -
    \\

    BAC
    & 17.3 & 81.3
    & 13.0 & 38.7
    & 8.4 & 21.9
    & - & -
    \\

    \midrule

    UAD (ours)
    & 11.4 & 19.7
    & 11.1 & 19.4
    & 8.5 & 19.8
    & 15.2  & 22.3
    \\

    \bottomrule

\end{tabular*}

\end{table}

\subsection{Blocking Latency of DiT}

VLA inference latency consists of two main components: a largely fixed context-processing overhead, dominated by the VLM, and the iterative DiT computation for action generation.
The average action availability latency (AAL) reported in Section \ref{sec:eval}
measures the end-to-end waiting time for actions and therefore accounts for both components.
However, UAD and most of the compared baselines primarily optimize the DiT computation, while leaving the VLM and other fixed overheads largely unchanged.
We therefore additionally report a DiT-specific latency metric to more directly characterize the reduction in denoising time that blocks action execution.
Specifically, we define \textbf{DiT blocking latency (DBL)} as the average amount of DiT computation per VLA inference that directly blocks action execution.
DiT computation that is successfully overlapped with physical action execution does not contribute to DBL.

Table~\ref{tab:aal_dbl} reports both AAL and DBL across the simulation settings and our real-world deployment.
UAD substantially reduces DBL in all evaluated settings.
Notably, the relative reduction in DBL is consistently larger than that in AAL.
This is expected because AAL additionally includes VLM processing and other fixed overheads that UAD, as well as most competing methods, does not optimize.
Consequently, DBL more directly exposes the reduction in action-blocking DiT time, further demonstrating the effectiveness of UAD in reducing action waiting time.

\subsection{Meta-World+ Difficulty Breakdown}

\begin{table}[t]
\centering
\caption{
    Success rate (\%) on Meta-World+ by task difficulty
    (easy / medium / hard / very hard: 28 / 11 / 6 / 5 tasks, 20 episodes each).
}
\label{tab:mw_difficulty}

\footnotesize
\setlength{\tabcolsep}{1.8pt}
\renewcommand{\arraystretch}{1.08}

\begin{tabular*}{\linewidth}{
    @{\extracolsep{\fill}}
    lcccccccccc
    @{}
}

    \toprule

    &
    \multicolumn{5}{c}{\textbf{SmolVLA}}
    &
    \multicolumn{5}{c}{\textbf{$\bm{\pi_{0.5}}$}}
    \\
    \cmidrule(lr){2-6}
    \cmidrule(lr){7-11}

    Method
    & Easy
    & Medium
    & Hard
    & Very hard
    & Overall
    & Easy
    & Medium
    & Hard
    & Very hard
    & Overall
    \\

    \midrule

    Vanilla ($N^*$)
    & 85.2 & 56.4 & 50.0 & 47.0 & 70.8
    & 80.7 & 58.6 & 68.3 & 42.0 & 70.5
    \\

    Vanilla ($N=1$)
    & 76.1 & 45.5 & 38.3 & 29.0 & 60.1
    & 79.8 & 54.1 & 55.0 & 34.0 & 66.6
    \\

    \midrule

    D3P
    & 84.1 & 57.7 & 50.8 & 44.0 & 70.3
    & 80.2 & 61.8 & 68.3 & 38.0 & 70.5
    \\

    RTC
    & 63.2 & 47.7 & 30.8 & 27.0 & 52.3
    & 69.8 & 52.3 & 49.2 & 33.0 & 59.8
    \\

    BAC
    & 80.4 & 45.9 & 38.3 & 31.0 & 62.8
    & 75.9 & 39.5 & 44.2 & 20.0 & 58.5
    \\

    \midrule

    UAD (ours)
    & 80.4 & 55.5 & 51.7 & 45.0 & 67.9
    & 80.0 & 58.2 & 67.5 & 35.0 & 69.2
    \\

    \bottomrule

\end{tabular*}

\end{table}

To further examine whether acceleration affects tasks of different difficulty differently, following the official breakdown, we separate the Meta-World+ tasks into difficulties of easy, medium, hard, and very hard, and present the results in Table~\ref{tab:mw_difficulty}.
Overall, the performance degradation introduced by aggressive acceleration tends to become more pronounced on more difficult tasks.
For example, uniformly reducing denoising to $N=1$, as well as RTC and BAC, causes substantially larger losses on the hard and very-hard subsets than on easy tasks in many cases.
This suggests that challenging tasks are generally more sensitive to degradation in action quality.
In contrast, UAD maintains competitive success rates across difficulty levels.
These results indicate that the favorable success-latency trade-off of UAD extends to challenging tasks and is not driven solely by performance on any difficulty group.

\subsection{Real-World Task Breakdown}

\begin{table}[t]
\centering
\caption{
    Success rate (\%) on individual real-world manipulation tasks.
    Each task is evaluated over 20 episodes.
    Detailed task descriptions are provided in Fig.~\ref{fig:real_task_gallery}.
}
\label{tab:real_world_breakdown}

\footnotesize
\setlength{\tabcolsep}{5.0pt}
\renewcommand{\arraystretch}{1.08}

\begin{tabular*}{\linewidth}{
    @{\extracolsep{\fill}}
    lcccccc
    @{}
}

    \toprule

    Method
    & Button
    & Grapes
    & Flower
    & Tidy
    & Stack
    & Overall
    \\

    \midrule

    Vanilla ($N=10$)
    & 100
    & 85
    & 65
    & 95
    & 75
    & 84
    \\

    Vanilla ($N=1$)
    & 100
    & 70
    & 55
    & 75
    & 55
    & 71
    \\

    \midrule

    UAD (ours)
    & 100
    & 90
    & 65
    & 90
    & 75
    & 84
    \\

    \bottomrule

\end{tabular*}

\end{table}

Our real-world evaluation consists of five manipulation tasks, and Table~\ref{tab:real_world_breakdown} reports the success rate for each task separately. Uniformly reducing the denoising budget to $N=1$ degrades performance on four of the five tasks, reducing the overall success rate from $84\%$ to $71\%$. In contrast, UAD maintains an overall success rate of $84\%$ and remains close to Vanilla ($N=10$) across all five tasks, indicating consistent performance preservation across different real-world manipulation behaviors.

\subsection{Evaluation on Additional Model-Benchmark Settings}
\label{app:sub_additional_settings}
\begin{figure}[t]
    \centering
    \includegraphics[width=\linewidth]{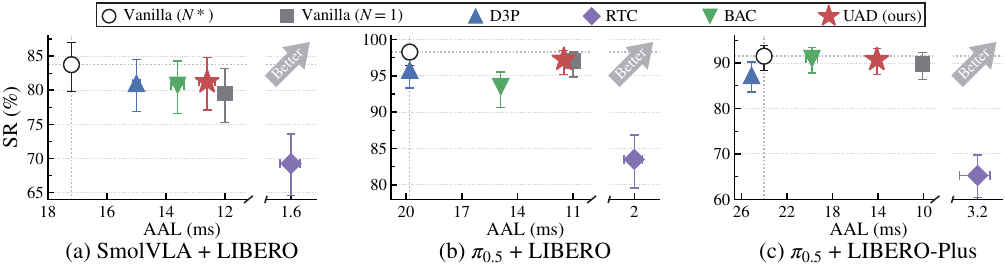}
    \caption{
        Success rate-latency trade-offs across three additional simulation settings.
        Broken horizontal axes accommodate RTC's substantially lower AAL. Error bars indicate Wilson 95\% CIs for SR, and episode bootstrap 95\% CIs for AAL. Some AAL CIs are too narrow to be visible at the plotted scale.
    }
    \label{fig:additional_settings}
\end{figure}

We extend our comparison to the three settings not included in the main evaluation: SmolVLA on LIBERO, and $\pi_{0.5}$ on LIBERO and LIBERO-Plus, using the same evaluation protocol.
As characterized in Appendix~\ref{app:denoising_budget}, these settings exhibit relatively small success rate differences between Vanilla ($N=1$) and Vanilla ($N^*$), making uniform single-step denoising a simple and effective acceleration strategy.

As shown in Fig.~\ref{fig:additional_settings}, Vanilla ($N=1$) achieves lower AAL than UAD with only modestly lower success rates, leaving limited room for UAD to improve upon uniform step reduction in these settings.
Nevertheless, UAD consistently reduces AAL relative to Vanilla ($N^*$) while maintaining comparable success rates.
Compared with D3P and BAC, UAD achieves lower AAL with comparable or higher success rates.
RTC attains substantially lower AAL but incurs pronounced success rate losses in all three settings.
These results complement the main evaluation: UAD remains effective across the additional settings, while its advantage over uniform step reduction is more pronounced when task success depends more strongly on the denoising budget.

\end{document}